\documentclass[lefttitle,review]{elsarticle}
\usepackage[ margin=1.5in]{geometry}
\usepackage{lineno}
\usepackage{natbib}
\usepackage{tabularx}
\usepackage{amsmath}
\usepackage{setspace}
\usepackage{amsfonts}
\usepackage{latexsym}
\usepackage{fancyhdr}
\usepackage{lastpage}
\usepackage{amsmath}
\usepackage{wrapfig}
\usepackage{color}

\makeatletter
\def\ps@pprintTitle{%
	\let\@oddhead\@empty
	\let\@evenhead\@empty
	\def\@rfoot{\thepage}%
	\let\@evenfoot\@oddfoot}
\makeatother

\biboptions{authoryear}

\usepackage{acronym}

\begin{document}
\begin{frontmatter}
\title{A Novel Deep Neural Network Architecture for Real-Time Water Demand Forecasting}
\author[1,2]{Tony Salloom}
\ead{Tonysalloom@ieee.org}

\author[1,2,3]{Okyay Kaynak}
\ead{Okyay.kaynak@boun.edu.tr}

\author[1,2]{Wei He\corref{mycorrespondingauthor}}
\cortext[mycorrespondingauthor]{Corresponding author}
\ead{Weihe@ieee.org}

\address[1]{School of Automation and Electrical Engineering, University of Science and Technology Beijing, Beijing 100083, China}
\address[2]{Institute of Artificial Intelligence, University of Science and Technology Beijing, Beijing 100083, China}
\address[3]{Bogazici University, Istanbul Turkey}

\begin{abstract}
Short-term water demand forecasting (StWDF) is the foundation stone in the derivation of an optimal plan for controlling water supply systems. Deep learning (DL) approaches provide the most accurate solutions for this purpose. However, they suffer from complexity problem due to the massive number of parameters, in addition to the high forecasting error at the extreme points. In this work, an effective method to alleviate the error at these points is proposed. It is based on extending the data by inserting virtual data within the actual data to relieve the nonlinearity around them. To our knowledge, this is the first work that considers the problem related to the extreme points.
Moreover, the water demand forecasting model proposed in this work is a novel DL model with relatively low complexity. The basic model uses the gated recurrent unit (GRU) to handle the sequential relationship in the historical demand data, while an unsupervised classification method, \textit{k}-means, is introduced for the creation of new features to enhance the prediction accuracy with less number of parameters. Real data obtained from two different water plants in China are used to train and verify the model proposed. The prediction results and the comparison with the state-of-the-art illustrate that the method proposed reduces the complexity of the model six times of what achieved in the literature while conserving the same accuracy. Furthermore, it is found that extending the data set significantly reduces the error by about 30\%. However, it increases the training time.
\end{abstract}

\begin{keyword}
Deep learning; Neural networks; Gated recurrent unit; Unsupervised classification;  Water demand forecasting; Extreme points
\end{keyword}
\end{frontmatter}
\doublespacing
\newpage
\section*{Abbreviations}
\begin{acronym}[SARIMASS]
	\acro{AIC}[AIC]{Akaike Information Criterion}
	\acro{ANN}[ANN]{ Artificial Neural network}
	\acro{ARIMA}[ARIMA]{ Auto-regression integrated moving average }
	\acro{BGRU}BGRU{Basic GRU-based model}
	\acro{CPU}[CPU]{Central processing unit}
	\acro{DCGRU}[DCGRU]{Data classification gated recurrent unit-based model}
	\acro{DL}[DL]{Deep learning}
	\acro{DMA}[DMA]{District metering area}
	\acro{DWT}[DWT]{Discrete wavelet transform}
	\acro{EDCGRU}[EDCGRU]{Extended data classification gated recurrent unit model}
	\acro{ELM}[ELM]{Extreme learning machine}
	\acro{GRU}[GRU]{Gated recurrent unit}
	\acro{LSTM}[LSTM]{Long short-term memory}
	\acro{MAE}[MAE]{Mean absolute error }
	\acro{MAPE}[MAPE]{Mean absolute persentage error }
	\acro{PCA}[PCA]{Principal component analysis}
	\acro{RAM}[RAM]{Random access memory}
	\acro{SARIMA}[SARIMA]{Seasonal auto-regression integrated moving average  }
	\acro{StWDF}[StWDF]{Short-term water demand forecasting}
	\acro{SVM}[SVM]{Support vector machine }
	\acro{Tanh}[Tanh]{Hyperbolic tangent function }
\end{acronym}

\section*{List of symbols}
\begin{acronym}[SARIMASS]
	\acro{b}[$ b $]{Bias matrices}
	\acro{Cj}[$ C_j $]{Center of class $ j $}
	\acro{i}[$ i $]{The index of the data samples}
	\acro{j}[$ j $]{The index of the data class}
	\acro{k}[$ k $]{Total number of the prameters of the model}
	\acro{m}[$ m $]{The number of classes}
	\acro{n}[$ n $]{Total number of data samples}
	\acro{p}[$ \rho $]{Numer of vertual values between two actual values}
	\acro{RSS}[$ RSS $]{Sum of square error}
	\acro{sigh}[$ \sigma_h $]{Inner activation function of GRU cell}
	\acro{sigg}[$\sigma_g$]{Output activation function of GRU cell}
	\acro{t}[$ t $]{The index of the prediction period }
	\acro{u}[$ U $]{Weight matrices of inner state of GRU}
	\acro{vt}[$ V_t $]{Water demand value during period $ t $ }
	\acro{w}[$ W $]{Weight matrices of input of GRU}
	\acro{wij}[$ W^{(i,j)} $]{The value of class $ j $ corresponding to sample $ i $ of the data}
	\acro{xi}[$ x_i $]{The sample $ i $ of data}
\end{acronym}

\section{Introduction}
Water scarcity has become a threat to humankind in recent decades. Many efforts in all possible directions are being made to compensate for this growing problem \citep{Northey2016, Gonzalez-Zeas2019}. The major reliable strategies for that include water treatment \citep{ZINATLOOAJABSHIR2020struc}, water desalination, and optimization of water management systems. Nanotechnology is the most powerful technology employed for water treatment, where researchers have done impressive work \citep{ZINATLOOAJABSHIR2020maryam,ZINATLOOAJABSHIR2017photo,moshtaghi2016preparation}. On the other hand,
StWDF is the foundation stone of the optimization of water management systems. Therefore, numerous researchers have directed their efforts towards this problem.
Nowadays, deep learning (DL) is the most dominant approach, which provides the most promising solutions to a myriad of critical problems. To mention a few examples, \cite{Zhou2020} has proposed deep learning method for forecasting the water quality in the presence of missing data.  \cite{Friedel2020} have compared four machine learning methods to predict groundwater redox status in the agriculturally dominated regions of New Zealand. \cite{Jiwen2020}  have used a deep belief network to forecast the depth of snow over in Alaska. Flood prediction is another crucial problem that utilizes deep learning approaches. An Encoder-Decoder based on Long Short-term-memory (LSTM) is proposed by \cite{kao2020exploring} for multi-step-ahead flood forecasting. While flood susceptibility modeling using deep learning is investigated in many studies, such as \citep{pham2021can, bui2020verification}. LSTM is also used by \citep{ni2020streamflow} for streamflow and rainfall prediction where two LSTM-based models are built, one combined wavelet network with LSTM and the other combined convolutional network with LSTM to achieve better performance.

StWDF is one of these problems that benefit most from DL to develop effective methods. However, some challenges might impede the success of DL-based solutions. Model complexity, the accumulative error when forecasting multi-steps, and the significant prediction error at the extreme points are some of these challenges that still need investigation. Model complexity and the size of the model in particular, become serious constraints when using federated learning approach, and bring some extra challenges to knowledge transformation technologies. Model complexity includes two types, time complexity and space complexity. Time complexity is brought on by the time required to train the model and by the data size required for training. The place complexity problem usually comes to the surface when the model has a massive number of parameters, which increases the model size, as well as the training time.

The second challenge is the accumulative error problem, which affects the multi-step prediction of water demand. When relying on the historical data of water demand for StWDF, the predicted values are involved in predicting the following values. Thus, the prediction error is compounded by the use of inaccurate values of the predicted water demand.

The third challenge is the significant error at the extreme points, which occur as a normal reflection of the nonlinearity of daily water demand. These points are recognized as periods where water demand is dramatically different from the average demand in the adjacent periods, making it difficult for the model to approximate the demand at these points. A few examples of extreme points are illustrated in Fig. \ref{fig6} and Fig. \ref{fig7}. The error at these points is usually unacceptable and may lead to severe problems in the distribution system.

In literature, complexity problem was not a critical issue when statistical methods such as Auto-regression integrated moving average (ARIMA) method and the seasonal version of it (SARIMA), in addition to Markov Chain, are used for StWDF \citep{Pandey2021}. However, their accuracy is not sufficiently satisfactory. \cite{Caiado2010} has achieved $ 11\% $ of mean square percentage error for one-day prediction and $ 13.2\% $ for $ 7 $-days ahead prediction by combining SARIMA with the generalized autoregressive conditional heteroscedasticity method for daily WDF.
\cite{Arandia2016} also have achieved $ 4.21\% $ of mean absolute percentage error (MAPE) for 15-minutes prediction of water demand in Dublin Spain by involving some data assimilation technique to improve the performance of SRIMA method.
However, their proposal does not work well with hourly prediction, where the best MAPE they have got is $ 38.12\% $. \cite{Brentan2017} have built their prediction model based on  SVR and Fourier methods for on-line prediction of hourly water demand. Their model achieves MAPE of $ 3.41\% $ for one-step prediction.

With the expansion in the application of machine learning \citep{zhou2020data,he2019disturbance} and artificial neural networks (ANNs) \citep{Salloom2019,yu2020sea,he2017survey}, several studies have proved that machine learning methods outdo the stochastic and the probabilistic models for WDF. \cite{Gagliardi2017}  has proved the ANNs overcome Markov Chain-based models in terms of forecasting accuracy.
\cite{Herrera2010a} and \cite{Bai2015} have proved the efficiency of support vector regression method for hourly WDF.
\cite{Guancheng2018} has compared statistical methods and conventional ANNs with deep learning method and proved that the DL methods give more accurate results when predicting water demand for short-horizon.

In fact, most researchers focus on improving prediction accuracy without paying too much attention to model complexity. Furthermore, a new trend that exacerbates this problem has started looming on the horizon recently, where some researchers comprise many machine-learning models in one system. Then, they chose a different one for different prediction periods based on probabilistic methods. \cite{Ambrosio2019} have used a combination of multilayer perceptron, SVM, ELM, random forests and adaptive neural fuzzy inference systems for hourly WDF.
\cite{Antunes2018} have investigated combining SVR, ANN, $ k $-nearest-neighbours, and random forest regression for real-time WDF. This strategy is meant to use the most accurate model in the most suitable prediction period. However, it requires a massive amount of computations and memory to save the parameters of all models and system configuration setting.

The error at extreme points also contributes to the worsening of the prediction accuracy; however, It has not attracted researchers' attention. \cite{Guancheng2018} are the first to point out this problem in their work, but they have not provided any solution. 

The accumulative error problem may occur when predicting several steps ahead based on the historical data of water demand. Some researchers tried to solve this problem individually by building a new neural network model and train it to make the predicted values approach the real ones. Obviously, this method increases the parameter of the system dramatically.

In fact, the accumulative error problem can be mitigated by involving several factors besides the historical data \citep{Papageorgiou2015,Holsteg2020}, so the impact of the predicted values in the input can be reduced efficiently.
Many factors influence water demand level \citep{Dias2018}, but only factors such as meteorological conditions and day type, which have weekly or daily distinguishable changes, have real impacts on StWDF \citep{Romano2014}. However, managers still rely on the historical data of water demand for StWDF, and ignore the other possible factors because of the difficulties of gathering data about them in real time, particularly in the 15-minutes interval. Moreover, some available information about some factors, such as meteorological information, are not sufficiently accurate for short time prediction \citep{Rayner2005}, making them unreliable.

In this work, we propose a novel DL model for StWDF. It is built based on the gated recurrent unit (GRU), and unsupervised classification method $k$-means. Involving data classification as a prior step has two major benefits, (i) it helps with creating new features to compensate for the leak of reliable features, which reflects positively on the prediction accuracy and the accumulative error. (ii) It creates a relationship between data of different days, which an ANN can be easily approximated with a small number of parameters, which in turn enhances the space complexity of the models.

Additionally, we investigate using a novel technique to alleviate the nonlinearity at the extreme points and reduce the error. It depends on inserting virtual data between the actual data so that the nonlinearity at these points is drastically declined.

The contribution of this paper can be summarized as follows: 
\begin{itemize}
	\item A novel DL model that provides a high prediction accuracy for both one step (15-minutes) and multi-step (96 steps ahead) prediction is proposed. It is built based on GRU neural network, supported by an unsupervised classification step to enhance the accuracy.
	\item A new technique for mitigating the prediction error at extreme points is proposed and investigated.
	\item The accumulative error problem, which occurs in multi-step prediction, is mitigated by means of classification step, which establishes new features to rely on in the prediction. 
	\item A comparison with the state-of-the-art is carried out to show the effectiveness of our proposed methods in terms of accuracy and model complexity.
\end{itemize}

The rest of this paper is organized as follows: Section \ref{Sec2} describe the equipment used to carry out this research. Section \ref{Sec3} explains the research methodology and the methods proposed in this work. Section \ref{Sec4} clarifies the results of this research, including the structure of the prediction model and the evaluation results. While Section \ref{Sec5} provide an intensive discussion to illustrate the underlying cause of these results. Section \ref{Sec6} illustrates the significance of the results.  Section \ref{SecConc} includes the conclusion and the planned future works.
\section{Research equipment and tools }
\label{Sec2}
The machine used to carry out this research including classification step, training the DL models, prediction process, and verification and evaluation of the proposed method is an ASUS laptop with an Intel Core i7 processor, four real cores, with a speed of 2.6 GHz each. The installed RAM is 16 GB. All models are built using Python 3.6 programming language over Anaconda platform.
Keras library and Tensorflow backend are used to build the DL models due to their availability and convenience, so anyone can simply redesign the models in a short time.
In fact, the proposed method for StWDF does not contain complicated steps that require very powerful hardware.
Ten years of water demand data do not exceed 1GB if saved in a CSV file, which means that small storage space is sufficient to store the required data.
All calculations can be achieved using a CPU with the specification mentioned above as the minimum requirement. The time spent on prediction and training using the aforementioned specifications are listed in the results' Section. Thus the proposed model is achievable in the industrial field by the currently available model.

\section{Methodology}
\label{Sec3}
Firstly, the water demand is collected every 15 minutes from the water distributing system, and the database is updated every 24 hours. One-step prediction scenario and multi-step prediction scenario are considered as described in Section \ref{Sub2Sec3}.
$ k $-means method is applied to classify the data based on their numerical distance into $ m $ classes as described in Section \ref{Sub3Sec3}. The number $ m $ is determined using Elbow method. Then, demand readings and the classes are organized in vectors, $ (V_t, C_1, C_2,\dots, C_m) $, each vector contains the demand reading $ V_t $ in addition to the value of each class that related to the corresponding demand value ($ C_1, C_2,\dots, C_m $), where the value of the class ($ C_j $) is determined as in equation \ref{Eq5}. The last 96 values of water demand are used to predict the demand in the following 15 minutes. In Scenario 1, prediction is made by passing 96 vectors to the dense block. This block handles each vector individually and results in one value for each vector. Thus, the output of the dense block contains 96 values. Next, these 96 values are passed to the GRU block. The output of the GRU block is the desired demand value.  In Scenario 2, the prediction is achieved iteratively by repeating Scenario 1 for 96 times. The result of each iteration is used in the next iteration.

The prediction model consists of the dense block and the GRU block, as explained in Section \ref{Sub4Sec3}. The final structure of the prediction model is clarified in Section \ref{Sub1Sec4} and shown in Fig. \ref{fig2}. the model is trained on the training data set as described in Section \ref{Sub6Sec3}.

In order to solve the problem of the massive error at the extreme points, the data set is extended by injecting virtual values between each two adjacent actual values such that the linearity is increased. The proposed method is described in Section \ref{Sub5Sec3}.

The accuracy of the proposed methods is evaluated based on the mean absolute error (MAE) and the mean absolute percentage error (MAPE). The complexity of the model is assessed based on the Akaike Information Criterion (AIC). The complete evaluation methodology is described in Section \ref{Sub8Sec3}.

\subsection{Water demand data}
\label{Sub1Sec3}

\begin{figure}[h!]
	\centering
	\includegraphics[scale=1]{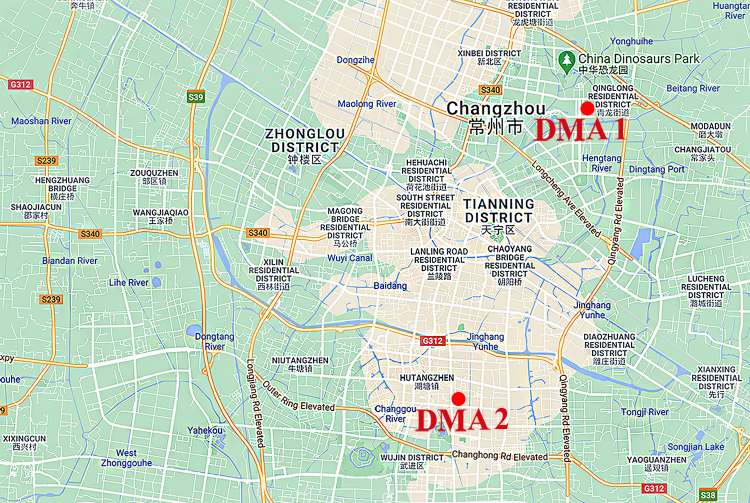}
	\caption{The locations of DMA1 and DMA2 on the map.}
	\label{figMAp}
\end{figure}
The data used in this research are collected in 2016, from two different district metering areas (DMAs) in Changzhou city in China \citep{Guancheng2018}. Fig. 1 shows the location of the two DMAs on the map as two red spots. DMA1 is a residential area with  about 13000 residences and a few commercial buildings, while the second area DMA2 is an industrial area with a population of about 8500 and 300 factories. The data set contains 25000 measured demand values for each DMA. Each day is divided into 96 duration, each of 15 minutes, and the total water demand is measured for the whole DMA at the end of each duration. 

The time recorded in the database is the end of each duration. The database is updated every 24 hours, i.e. the actual values of water demand for each day are not available until the end of the day.
Table \ref{DataInfo} lists the statistical information of the data used in this research.

\begin{table}[!t]
	\begin{center}
		\caption[0.7\linewidth]{Statistical information of water demand data}
		\label{DataInfo}
		\begin{tabular*}{\linewidth}{ p{7cm}  p{2cm} p{2cm} }
			\hline
			Statistics  & DMA1 & DMA2\\
			\hline
			Number of readings  & $ 25000 $ & $ 25000 $\\
			Minimum demand ($m^3/15 $ min) & $ 30.0 $ &	$ 40.0 $\\
			Quartile(0.25)&	$ 56 $ &	$ 64 $\\
			Median ($m^3/15 $ min) &	$ 86 $ &	$ 89 $\\
			Quartile(0.75)&	$ 94 $ &	$ 104 $\\
			Maximum demand ($m^3/15$ min)	& $ 157.0 $	& $ 149.0 $\\
			Mean demand ($m^3/15$ min)  & $ 81.4 $ & $ 87.0 $\\
			Standard deviation&	$ 24.4 $ &	$ 21.1 $\\
			Mode ($m^3/15 $ min) & $ 92 $ &	$ 92 $\\
			Skewness ($m^3/15 $ min) & $ -0.478 $ &	$ -0.116 $\\
			kurtosis  ($m^3/15 $ min) & $ -1.07 $ &	$ -1.23 $\\			
			Type of DMA  &	Residential	& Industrial\\
			\hline
			m = meter, min = minute.
		\end{tabular*}	
	\end{center}	
\end{table}

Data set is divided into two sets; Training set and testing set. The training set is used to train the models.
It contains 22500 records. During training, 15\% of the training data is used for validation.
The testing set contains 2500 records used to test the accuracy of the models after training ends.
\subsection{Prediction scenarios}
\label{Sub2Sec3}
This work targets the problem of short-term water demand forecasting based on the historical data of water demand. Each prediction step is done for one prediction period, which is 15 minutes in length. Water demand values of the last 96 periods (i.e. the demands of the previous 24 hours) are employed to achieve each prediction step.

Two forecasting scenarios are considered; (i)Scenario 1 is a one-step forecasting scenario; where the actual values of the required historical data are assumed to be available when the prediction starts. (ii)Scenario 2 is a multi-step forecasting scenario, where 96 steps, each one is of 15 minutes, are forecasted iteratively. In this scenario, the actual values are available only for the first prediction step, while the required data for the following steps contain recently predicted values in addition to the available previous actual values when they are available. In fact, this is the realistic scenario since the database considered in this research will be updated at the end of each day.
\subsection{Data classification and feature building}
\label{Sub3Sec3}
The essential step of building an ANN model for any purpose is identifying the inputs to that model. 
In this research, the historical water demand data is the only available data for prediction. 
we choose to use the demand values of the last 96 periods ($ t-1,t-2, \dots, t-96$) for StWDF for each prediction period ($ t $).

In order to compensate for the lack of features, the available data is classified into $ m $ classes. Then, the created classes are used as new features. Since there is no clue about the possible features, to simplify the method, a simple version of the algorithm $k$-means is applied to implement an unsupervised classification of the historical data of water demand based on the numerical distance between values. The number $ m $ should be identified carefully, due to the fact that large $ m $ undermines the benefit of classification. On the other hand, small $ m $ increases the within-cluster square error, also called Distortion \citep{yang2019}. Elbow Method is used widely in the literature and in this research to determine the optimal number of classes to be used. According to this method, the best value of $ m $ is the smallest number that guarantees a small distortion. The distortion is calculated based on the following equation \citep{yang2019}:

\begin{equation} \label{Eq4}
SSE=\sum_{i=1}^{n} \sum_{j=1}^{m}w^{(i,j)}\Vert{x_i - C_j}\Vert^2_2
\end{equation}

Where $n$ is the total number of data samples, and $i$ is the sample index in the class $j$, $m$ is the total number of classes and $j $ is the class’s index. $w^{(i,j)}$ is calculated as follows:

\begin{equation}
\label{Eq5}
w^{(i,j)} =
\begin{cases}
1 \quad \quad  x_i \in Class_j \\  0 \quad \quad otherwise    
\end{cases}
\end{equation}

The initial center of each class $j$ is determined randomly as in the following formula:

\begin{equation}\label{Eq6}
Cj=randn(1,m)*Std+mean
\end{equation}

where $randn(a,m)$ generates an array of $(a \times m)$ items, each of them is a random float sampled from the normal distribution. $Std$ and $mean$ are the standard deviations and the mean of each data set, respectively. They are involved in (\ref{Eq6}) in order to guarantee that the created centers represent the whole data.
\subsection{DL model design}
\label{Sub4Sec3}
By classifying the data, new relationships between them are created, where data belongs to the same classes has a high level of similarity. To get the benefits of classification, the designed DL model should be able to approximate the relationship between each value and the classes. Additionally, the model should be able to approximate the sequential relationship between water demands. In this research, 96 records of the historical data are used to achieve every prediction step. 
Considering these requirements, the model comprises two blocks, the dense block that works upon the relationship of the water demand value with the classes, and the GRU block that works upon the sequential relationship between water demand data. Thus, the proposed model called data classification-based neural network model or DCGRU model in the rest of this paper.
\subsubsection{Dense Block}
\label{Sub1sub4Sec3}
The input of the model takes the form of a $ (96 \times (m+1) ) $ matrix, 96 is number of the rows where each row contains $ (m+1) $ values, which are one demand value and the corresponding values of the created classes. 
Then the input is passed into three fully connected layers of the type "time-distributed dense." Time-distributed dense layer is a dense layer that accepts two-dimensional input and applies the activation function to every row separately, and the output size equals the number of the rows in the input. Thus, the output of this part of the model is a vector of 96 values; each of them implicitly represents the water demand value of one record and its relationship to the created classes. 
When choosing the activation functions for layer 1 and layer 2, we consider getting a positive number at the output of each layer may accelerate the convergence. While an amplifier function is needed in the output layer.
The number of units in each dense layer is determined using grid search taking into account that maintaining a small number of units in each dense layer is preferred in order to arrive at a simple neural network model. 
\subsubsection{GRU Block}
\label{Sub2sub4Sec3}
The output of the dense block is a sequence of 96 values. It is passed into the GRU block, which contains a hidden GRU layer, and one GRU cell represents the output layer of the DCGRU model. The number of units in the hidden GRU layer is chosen to what suggested in the previous research \citep{Chung2014}. The input vector elements are passed to the GRU layer consecutively, one item at a time. The output of the GRU layer is calculated based on three activation functions, two inner functions, and one for the output. The sigmoid function is employed in several works in the literature \citep{Xu2019}, as an inner activation function for GRU layers, while the hyperbolic tangent (Tanh) is used as an output activation function. These three functions are applied to the current input and the output of the previous unit based on the following equations \citep{Deng2019}:

\begin{equation}\label{Eq1}
z_t=\sigma_g (W_z x_t+U_z h_{t-1}+ b_z )
\end{equation}
\begin{equation}\label{Eq2}
r_t=\sigma_g (W_r x_t+U_r h_{t-1}+ b_r)
\end{equation}
\begin{equation}\label{Eq3}
h_t=(1-z_t)*h_{t-1}+z_t *\sigma_h (W_h x_t+U_h (r_t * h_{t-1})+ b_h)
\end{equation} 
where $ x_t $ is the input vector, $ h_t $ is the output vector, $ z_t $ is the update gate vector, $ r_t $ is the reset gate vector, and $ W $, $ U $ and $ b $ are the trainable parameters matrices and vectors, while $ \sigma_g $ and $ \sigma_h $ are the sigmoid and the hyperbolic tangent functions, respectively.
The hyperparameters of the proposed model structure are listed in Table \ref{DCGRUInfo}.

\subsection{Expanding the data set}
\label{Sub5Sec3}
In order to alleviate the error at the extreme points, we suggest expanding the data set. In fact, the problem shows up because of the high difference between the water demand value at the extreme points and that at the points around them as shown in Fig. \ref{fig6} and \ref{fig7}. 
To solve this problem, we try to reduce this difference by setting a number $ \rho $ of virtual water demand values between every two consecutive actual values.
The new values are inserted between every two consecutive actual values. Next, the new values are classified based on the $k$-means method. 

In practice, the major problem of the method presented is the determination of the suitable $ \rho $ to be inserted into the data set.
At first, glance, increasing the number of virtual values seems to be suitable for reducing the error at the extreme points.
However, giving it more thought, the GRU layer remembers the sequential changes between input data, not data itself and reflects that on the weights of the model.
Thus, increasing the length of the linear sequence between two actual demand values reflects negatively on the next step of forecasting, which exacerbates the error.
Moreover, inserting many virtual values enlarges the input, which in turn enlarges the training and forecasting time. The value of $ \rho $ is determined experimentally taking into account the aforementioned thoughts.  

The input of the model is also expanded to comprise the actual water measurements and the virtual water values for a full day, i.e. The input size becomes $ (96(\rho + 1)) $.
The structure of the prediction model does not change. However, we will use the abbreviation EDCGRU to refer to the DCGRU model applied to the expanded data set in the rest of this article.

When predicting water demand for the next period in Scenario 1, all virtual values should be predicted one by one and classified, then included in the input data to achieve one prediction step. In Scenario 2, $ 96(\rho + 1) $ values of water demand should be predicted, consecutively. Every $ (\rho + 1) $ predicted value is considered a forecasting result of the corresponding forecasting period.
\subsection{Training}
\label{Sub6Sec3}
All models are trained under the same circumstances using the same equipment and the same data. The parameters of all models are initialized based on the Xavier uniform initializer. It initializes the weights with random numbers picked from a uniform distribution within an interval limited by values related to the number of input and output weights of the corresponding layer. 

The designed model is a compound of two blocks, Dense block, and GRU block. Each block uses different activation functions. Besides, each block acts upon different kinds of features. Therefore, a different trainable parameter of the model requires a different learning strategy. 
Adam training algorithm is used to train all models in this research.
It trains every weight in the neural network using different learning rates according to its previous changes, which match our learning strategy desired. Also, Adam algorithm distinguishes itself from the others because it is a quick trainer and memory conservative \citep{Sun2019}, which is one goal of this research. Mini-batch technique is used during the training. In this technique, the prediction error is calculated for a small batch of prediction steps, and then the average error is backpropagated to adjust models' weights and biases.

In order to guarantee that the training data are not delivered to the model in a meaningful order, the training data is shuffled after every epoch.

Although all models are trained similarly, however, different training parameters are required for a different model to result in good prediction accuracy. Table \ref{TRAINING_PARAMETERS} lists the training parameters for every model.
\subsubsection{Training parameters of the DCGRU model }
\label{Sub1Sub6Sec3}
For this model, the learning rate is chosen to what is recommended in \citep{kingma2014}. However, other values around 0.001 were examined and found that 0.002 is the best learning rate for DMA1, while 0.001 is the best learning rate for DMA2. $ \beta_1 $ and $ \beta_2 $  are set to 0.9 and 0.999, respectively, for both DMAs. In order to avoid overfitting, the early stop technique is used with a tolerance of two consecutive epochs, i.e. the training stops automatically when the prediction error of training data keep descending while the error of validation data starts ascending for two consecutive epochs. The final number of epochs is 15 and 18 for DMA1 and DMA2, respectively. The batch size is 100 for both DMAs.
\subsubsection{Training parameters of the EDCGRU model }
\label{Sub2Sub6Sec3}
By expanding the input, the initial values of the input layer changes, Thus, different values for the training parameter are required. The learning rate is scheduled to start at 0.002 and decreases by 50 per cent every 5 epochs for both DMAs. Epochs number is determined using the early stop technique with a tolerance of 4 epochs. The final epochs' numbers are 17 and 21 for DMA1 and DMA2, respectively. Batch size is set to 100.
\subsection{Comparison methodology}
To show the effectiveness of the proposed methods, we compare our model with two models. The first one is the basic forecasting model, referred to as "BGRU model". It is similar to the DCGRU model but it does not involve the classification step. Through this comparison, the benefits of the classification step to enhance the prediction accuracy is unveiled. The second one represents the state-of-the-art to show the superiority of the proposed model in terms of accuracy and space complexity. The GRUN model proposed by \cite{Guancheng2018} is chosen to compare with for the following reasons: authors present comprehensive research about the efficiency of the deep learning method and provide strong evidence about the superiority of deep learning over other possible methods. Thus, comparing the model proposed in this work with the GRUN model is sufficient to prove that our model is better than those in the state-of-the-art. They also consider the two forecasting scenarios that are considered in this research. Moreover, they use the same data sets we use for training, validation, and testing. The structures of these two models are illustrated in the Section \ref{Sub1Sub7Sec3} and Section \ref{Sub2Sub7Sec3}.

Additionally, To increase the scientific value of this work, a comparison with the results described in previous publications is listed in Table 7. the publications that consider SVM or neural network-based methods are included due to their superiority in StWDF over other methods. We compare this work with six works from the literature. A brief description of these works is listed here:
\begin{itemize}
	\item \cite{Du2021LSTM} combine discrete wavelet transform (DWT) and principal component analysis (PCA) with LSTM for one-step daily WDF.
	\item \cite{Antunes2018} use SVR, ANN, $ k $-nearest-neighbors, and random forest regression for one-step prediction of daily water demand. Then they choose the best result based on probabilistic methods.
	\item \cite{Brentan2017} combine SVM and Fourier method for one-step prediction of daily water demand.
	\item \cite{Mouatadid2017} use EML for one-step prediction of daily water demand.
	\item \cite{Candelieri2017CSVM} uses clustering to enhance SVM method for one-step prediction of hourly water demand.
	\item \cite{Tiwari2016} use ELM with wavelet ANN for one-step daily WDF.
\end{itemize} 
Readers can refer to the original works for more details.
\subsubsection{Basic GRU forecasting model (BGRU)}  
\label{Sub1Sub7Sec3}
\begin{figure}[!h]
	\centering
	\includegraphics[scale=0.7]{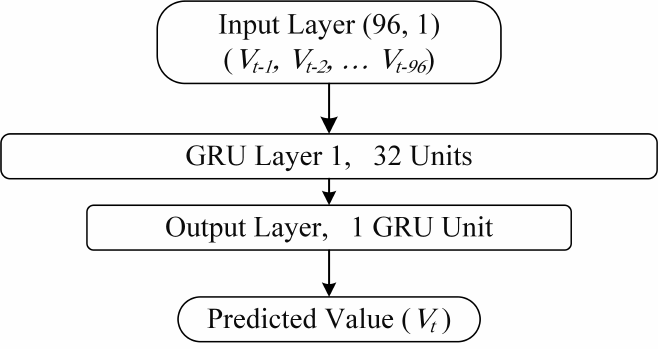}
	\caption{The structure of the BGRU model.}
	\label{fig3}
\end{figure}

BGRU model composed of the GRU block of the CDGRU model proposed in Section 3.4.2. The input of the BGRU model is a list of 96 items ($ V_{t-1}, V_{t-2}, \dots, V_{t-96} $), which represent water demand in the last 96 duration, ($ t-1, t-2, \dots , t-96 $), where (t) is the index of the duration for which the water demand is forecasted. It takes the shape of (96, 1). The hidden layer is a GRU layer with 32 units. The output layer is one GRU unit. The structure of this model is shown in fig.\ref{fig3}.

Adam training algorithm is used for training this model with a learning rate of 0.001 for both DMAs. Batch size is 100, and the final epochs numbers are 19 and 23 for DMA1 and DMA2, respectively. All other training parameters are similar to that of the DCGRU model listed in Table \ref{TRAINING_PARAMETERS}.

Prediction is made in the same way as when using the DCGRU model; however, in Scenario 2, the output of the current prediction step is included in the input of the next prediction step without any prior classification.
\subsubsection{GRUN forecasting model}  
\label{Sub2Sub7Sec3}
In this section, we provide a brief description of the method and the GRUN model structure. Readers can refer to the original work in \cite{Guancheng2018} for more details.
In this work, the authors extract three features out of these data based on the time: Recent Time, Near Time, and Distant Time. They select a small number of water demand measurements (5 values) for each feature.

\begin{table}[h!]
	\renewcommand{\arraystretch}{1.3}
	\caption{Selected water demand values for GRUN input}
	\label{GRUNFeatureInfo}
	\centering
	\begin{tabular*}{\linewidth}{p{4cm} p{8cm} }
		\hline
		Feature	& Selected demand values\\ 
		\hline
		\textit{Recent Time}	& $V_{t-1}, V_{t-2}, V_{t-3}, V_{t-4}$, and $V_{t-5}$ \\
		\textit{Near Time}	& $V_{t-94}, V_{t-95}, V_{t-96}, V_{t-97}$, and $V_{t-98}$ \\
		\textit{Distant Time} & $V_{t-190}, V_{t-191}, V_{t-192}, V_{t-193}$, and $V_{t-194}$ \\
		\hline
	\end{tabular*}
\end{table}

Table \ref{GRUNFeatureInfo} shows the selected demand values for each feature, where (t) is the desired time point where water demand should be predicted. The structure of the GRUN model includes three GRU layers, each layer acts upon one of the three features. The outputs of the three layers are merged and used as input to seven dense layers that used to discover the influence of each feature on the output. The researchers of the work changed the activation functions of the reset gate and update gate into ReLU function instead of sigmoid function. For forecasting Scenario 2, the forecasting accuracy decreases as the dependence on the predicted water demand values increases. In order to avoid that, the authors added a correction module to reduce the forecasting error, this module consists of one dense layer with 96 units. The parameters of the GRUN model and the correction model structure are listed in Table \ref{GRUNParameters}.

\begin{table}[h!]
	\caption{GRUN model structure}
	\label{GRUNParameters}
	\centering
	\begin{tabular*}{\linewidth}{p{4cm} p{5cm} p{4cm}}
		\hline
		Parameter &	GRUN Model & Correction  Model\\ 
		\hline
		\textit{Layers} &	3 GRU \newline  7 Dense &	1 \\
		\textit{Unites} &	$(48, 32, 32)$ GRU \newline  $(64, 32, 16, 8, 4, 2, 1)$ Dense &	$96$ Dense \\
		\textit{Activation function} & (Tanh, ReLU) GRU \newline (ReLU, Linear) Dense &	Linear\\
		\hline
	\end{tabular*}
\end{table}

The GRUN model is trained using Adam with the same data set described in Section \ref{Sub1Sec3} and the same equipment. All training parameters are listed in Table \ref{TRAINING_PARAMETERS}.

\begin{table}[h!]
	\
	\renewcommand{\arraystretch}{1.3}
	\caption{Training parameters for all prediction models}
	\label{TRAINING_PARAMETERS}
	\centering
	\begin{tabular*}{\linewidth}{p{2.2cm} p{2.5cm} p{2.5cm} p{2.5cm} p{2.5cm}}
		\hline
		Parameter &	BGRU & DCGRU & GRUN & EDCGRU\\ 
		\hline
		\textit{Initializer} & Xavier uniform & Xavier uniform & Xavier uniform & Xavier uniform \\
		\textit{Optimizer} & Adam & Adam & Adam & Adam\\
		\textit{Batch size} & 100 & 100 & 60 & 100\\
		\textit{Learning rate} & $0.001^{ a}$ \newline $0.001^{b}$ & $0.002^{ a}$ \newline $0.001^{b}$ & $0.002^{ a}$ \newline  $0.001^{b}$ & scheduled \\
		\textit{Epochs} & $19^{a}$ \newline $23^{b}$ & $15^{ a}$ \newline $18^{b}$ & $16^{a}$ \newline $20^{b}$ & $17^{a}$ \newline $21^{b}$ \\
		\hline
		\multicolumn{3}{l}{$^a = DMA1$ \newline $^b = DMA2$}\\
	\end{tabular*}
\end{table}
\subsection{Evaluation methodology}
\label{Sub8Sec3}
Each model is evaluated based on (i)Forecasting accuracy, (ii) the model space complexity, and (iii)computational load, including training and forecasting time.

(i)Forecasting accuracy is evaluated for individual forecasting scenarios based on the mean absolute error (MAE) and mean absolute percentage error (MAPE).
The testing data are used for evaluation.

(ii)The model complexity is calculated based on the Akaike Information Criterion (AIC). AIC is a reliable tool for selecting the best between several competing statistical models depends on the number of variables and output error of the models when applying the same data set. The model with the lowest AIC is deemed to be the best. AIC for a neural network model can be calculated by the following general equation \citep{Seghouane2011}:

\begin{equation} \label{Eq8}
AIC=n \ln{\bigg( \frac{RSS}{n} \bigg) } + 2k
\end{equation}

Where $n$ represents the number of observations, $k$ is the number of model’s variables, which are usually the weights and the biases of a neural network model, and $RSS$ represents the sum of square error of the model output.
When $ 1 < \frac{n}{k} < 40 $, then AIC needs a bias adjustment. In this case, the value of AIC can be calculated by the following equation \citep{Panchal2010a}:

\begin{equation} \label{Eq9}
AIC=n \ln{\bigg( \frac{RSS}{n} \bigg) } + 2k +\bigg( \frac{2k(k+1)}{n-k-1} \bigg)
\end{equation}
In this research, AIC is calculated using (\ref{Eq9}) based on the whole data where $n = 25000$ observations. The number of variables for each model is listed in Table \ref{EVALUATION_RESULTS}.

(iii)Training time is measured during training, where training data and validation data are employed. Forecasting time is measured for each prediction scenario individually using the testing data. In Scenario 1, time is measured for one-step prediction, while in Scenario 2, time is measured for 96-step prediction. The measurement is repeated 1000 times, and the average time is recorded.
\section{Results}
\label{Sec4}
\subsection{DCGRU model structure}
\label{Sub1Sec4}
\begin{figure}[!h]
	\centering
	\includegraphics[scale=0.7]{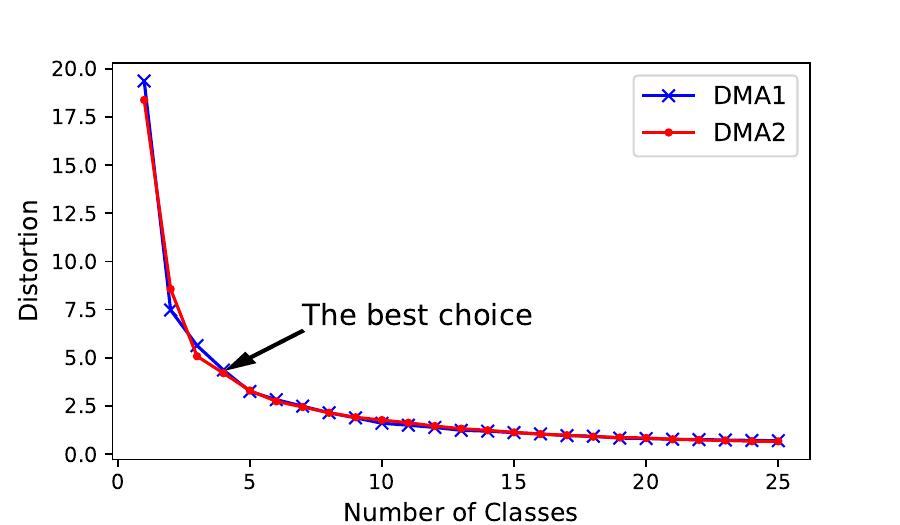}
	\caption{The Output of Elbow method.}
	\label{fig1}
\end{figure}

Fig. \ref{fig1} shows the distortion of water demand data in terms of the number of classes. It is clear that $4$ is the most suitable number of classes for both data sets of DMA1 and DMA2. 

\begin{figure}[!h]
	\centering
	\includegraphics[scale=0.8]{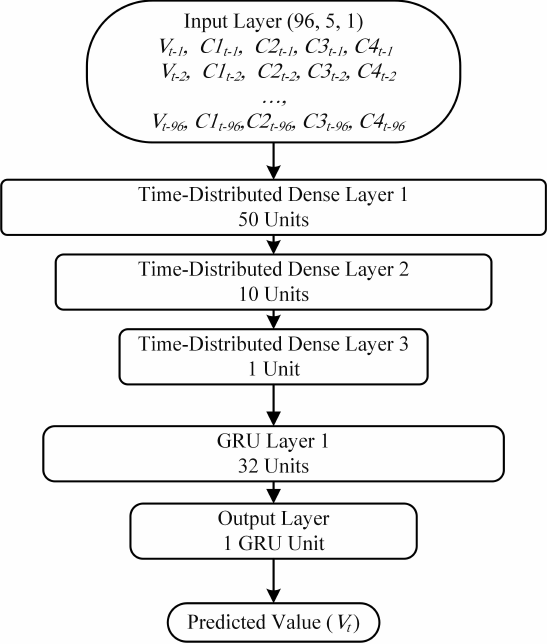}
	\caption{The structure of the DCGRU neural network model.}
	\label{fig2}
\end{figure}

Based on what described in the methodology Section \ref{Sub2Sec3}, the resulted model consists of two blocks as shown in fig.\ref{fig2}, (i)Dense block contains an input layer with a shape of (96, 5, 1) and two hidden layers. All layers are of type "Time-distributed dense." The input layer and the two hidden layers contain 50, 10, and 1 units, respectively. The activation function for both the input and the first hidden layer is ReLU function, while a linear function is used in the second hidden layer. (ii)GRU block contains two layers; GRU layer 1 contains 32 GRU cells; each cell has the sigmoid function for both of its inner activation functions, and the $ Tanh $ function for the output function. GRU layer 2 is the output layer of the DCGRU model. It contains only one GRU cell, with sigmoid and linear functions as its inner and output activation functions, respectively. Fig.\ref{fig2} illustrates the structure of the DCGRU model, and Table \ref{DCGRUInfo} lists the hyperparameters of this model.

\begin{table}[!h]
	\caption{The parameters of the DCGRU model}
	\label{DCGRUInfo}
	\centering
	\begin{tabular*}{\linewidth}{p{1.3cm} p{4cm} p{1.5cm} p{2.5cm} p{1.1cm}}
		\hline
		Layer &  Type of units & Number of units & Activation function & Output shape\\ 
		\hline
		$Input$ &  &  &  & (96,5,1)\\
		$1$ & Time-Distributed Dense & 50 & ReLU\\
		$2$ & Time-Distributed Dense & 10 & ReLU\\
		$3$ &	Time-Distributed Dense & 1 & Linear & (96, 1)\\
		$4$ & GRU & 32 & Sigmoid, Tanh\\
		$Output$ &  GRU & 1 & Sigmoid, Linear & (1, 1)\\
		\hline
	\end{tabular*}
\end{table}
After expanding the data, the input of the model changes based on $ \rho $. It is found that $ \rho = 1 $ enhances the prediction accuracy the most. Thus only one virtual value is added between every two consecutive actual values. In prediction Scenario 1, two consecutive values are predicted, the first one is the virtual value and the second is the actual value for this prediction step. In Scenario 2, 196 steps of prediction should be made, every second value is considered the actual value of the corresponding period.

\begin{figure}[h!]
	\centering
	\includegraphics[scale=0.60]{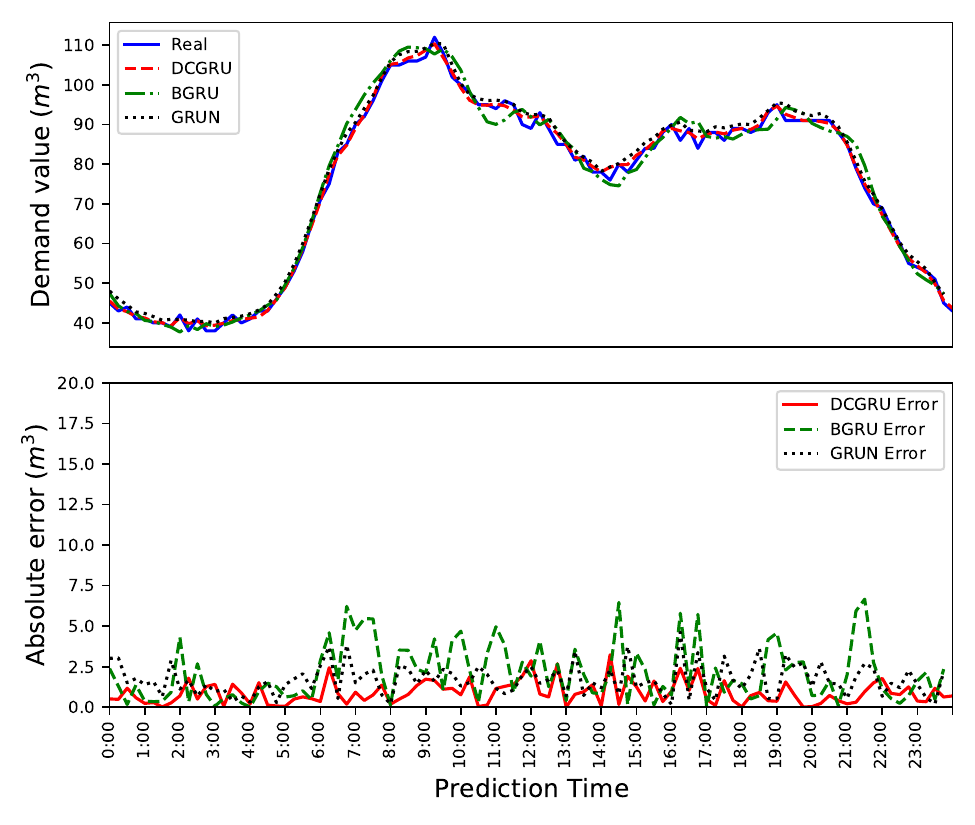}
	\caption{Prediction results and absolute errors for DMA1 in Scenario 1 without data expansion.}
	\label{fig4}
\end{figure}
\begin{figure}[h!]
	\centering
	\includegraphics[scale=0.60]{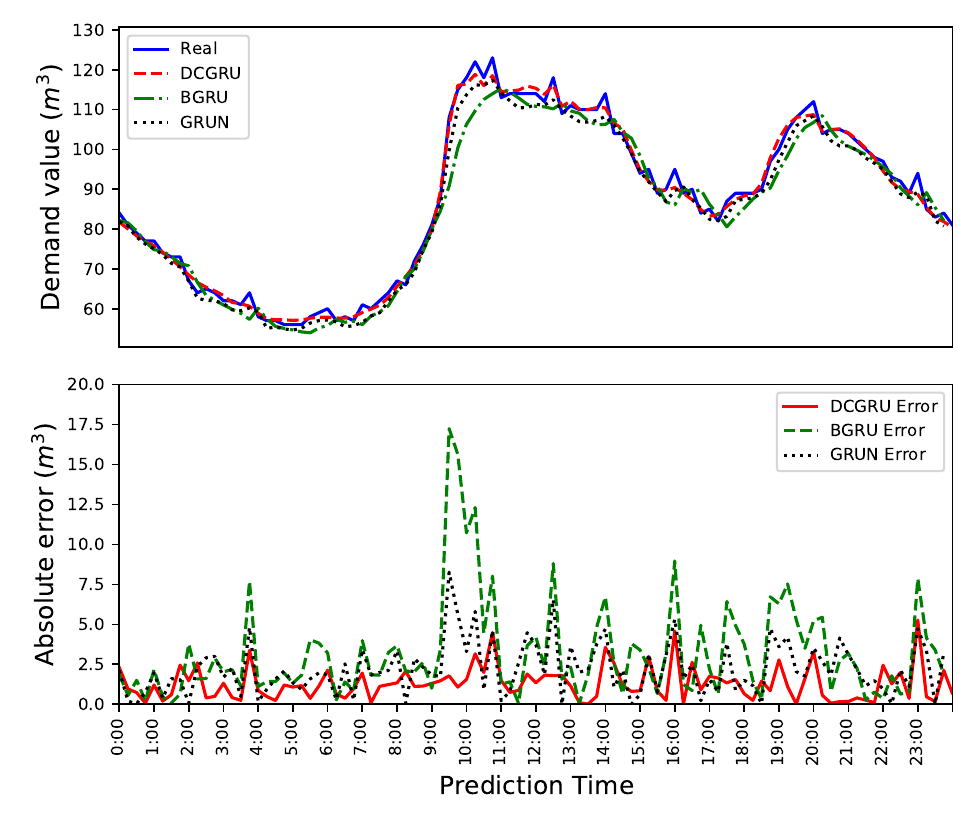}
	\caption{Prediction results and absolute errors for DMA2 in Scenario 1 without data expansion.}
	\label{fig5}
\end{figure}
\begin{figure}[h!]
	\centering
	\includegraphics[scale=0.60]{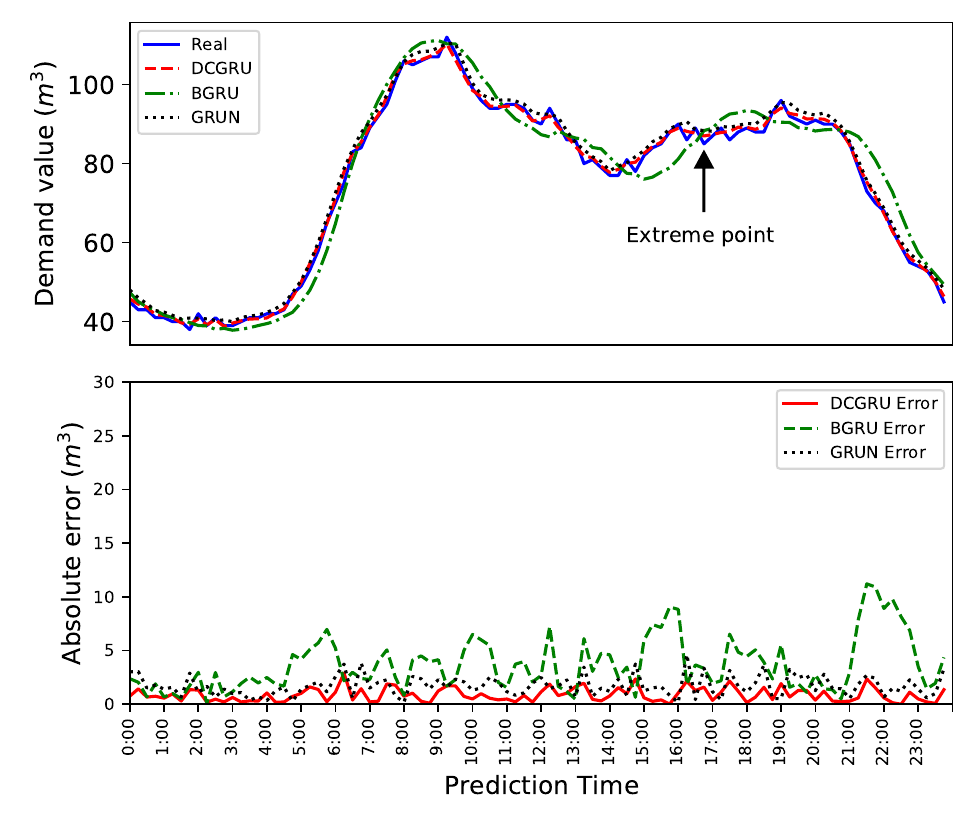}
	\caption{Prediction results and absolute errors for DMA1 in Scenario 2 without data expansion.}
	\label{fig6}
\end{figure}
\begin{figure}[h!]
	\centering
	\includegraphics[scale=0.60]{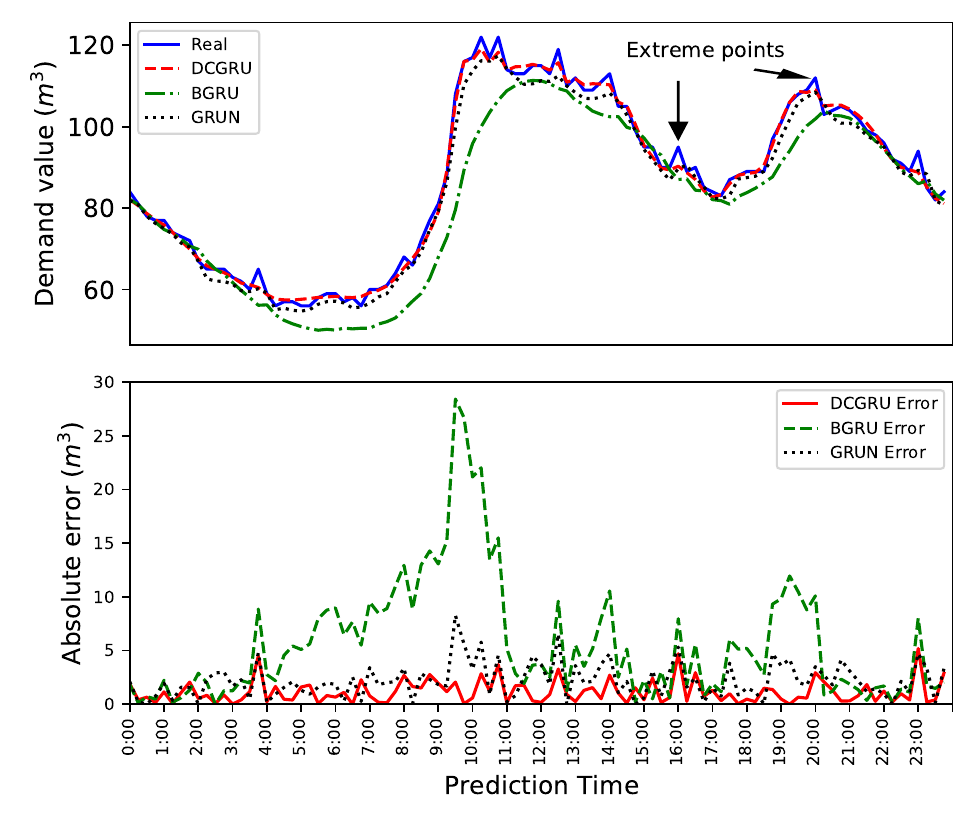}
	\caption{Prediction results and absolute errors for DMA2 in Scenario 2 without data expansion.}
	\label{fig7}
\end{figure}
\subsection{Prediction and comparison results}
\label{Sub2Sec4}
Fig.\ref{fig4} and Fig.\ref{fig5} illustrates a comparison between BGRU, DCGRU and the state-of-the-art model GRUN in terms of forecasting accuracy in Scenario 1 for DMA1 and DMA2, respectively. While Fig.\ref{fig6} and Fig.\ref{fig7} illustrate similar information in Scenario 2. Table\ref{EVALUATION_RESULTS} lists the numerical results of the comparison. The DCGRU model proposed in this work achieves the best forecasting accuracy between the three mentioned models. It achieves 1.26\% and 1.31\% of MAPE in Scenario 1 for DMA1 and DMA2, respectively, and 1.50\% and 1.38\% in Scenario 2 for DMA1 and DMA2, respectively. Table \ref{Previous_RESULTS} shows that our model achieves the best MAPE compared with the state-of-the-art. The MAPE rises to 2.49\% and 3.58\% when using the BGRU model for DMA1 and DMA2, respectively, in forecasting Scenario 1, and to 5.11\% and 7.03\% in Scenario 2. The GRUN model achieves a comparable accuracy in the state-of-the-art works, as illustrated in Table \ref{Previous_RESULTS}. Its MAPE reaches 2.06\% and 2.46\% in Scenario 1, and 4.33\% and 4.96\% in Scenario 2.
The compression with the results of the previous works, which listed in Table \ref{Previous_RESULTS}, shows that comparable accuracy is achieved by recurrent neural network-based methods, mainly LSTM and GRU neural network. The preprocessing step of the data makes a slight difference in their results. Guo et al., who did not add any preprocessing step, have achieved 2.06\% of MAPE, while Du et al., who used PCA and DWT to process the data before the prediction, have achieved 1.83\%.

The number of parameters $k$ of each model, which represents the number of weights and biases, depends on the structure of each model. Thus, DCGRU and EDCGRU have the same number of parameters which is 4187, while GRUN has 30403 parameters, and BGRU has 3366 parameters.

Before expanding the data set, The small number of parameters and the small error accomplished by the DCGRU model enhances the AIC for this model where it reaches 20633 and 29396 for DMA1 and DMA2, respectively, compared with 172225 and 176019 for the GRUN model. These numbers unveil the superiority of our model in terms of space complexity.

\begin{table}[h!]
	\renewcommand{\arraystretch}{1.6}
	\caption{Evaluation results}
	\label{EVALUATION_RESULTS}
	\begin{tabular*}{\linewidth}{p{4.2cm} p{1.5cm} p{1.5cm} p{1.5cm} p{1.5cm}}
		\hline
		Parameter &	BGRU & DCGRU & GRUN & EDCGRU\\ 
		\hline
		$k$ & $3366$ & $4187$ & $30403$ & $4187$\\
		$AIC$ & $51915^{a}$ \newline $74865^{b}$ & $20633^{ a}$\newline $29396^{b}$ & $172225^{ a}$\newline $176019^{b}$ & $9176^{a}$ \newline $11873^{b}$ \\
		\textit{S1 MAE} $ (m^3/15 min) $ & $1.91^{a}$\newline $3.22^{b}$ & $0.92^{a}$\newline $1.11^{b}$ & $1.58^{ a}$\newline $2.00^{b}$ & $0.59^{ a}$\newline $0.63^{b}$ \\
		\textit{S2 MAE} $ (m^3/15 min) $ & $3.69^{a}$\newline $5.96^{b}$ & $1.01^{ a}$\newline $1.17^{b}$ & $3.64^{ a}$\newline $4.03^{b}$ & $0.72^{a}$\newline $0.73^{b}$ \\
		\textit{S1 MAPE (\%)}& $2.49^{a}$\newline $3.58 ^{b}$ & $1.26 ^{ a}$\newline $1.31 ^{b}$ & $2.06 ^{ a}$\newline $2.46 ^{b}$ & $0.98 ^{a}$\newline $0.99^{b}$ \\
		\textit{S2 MAPE (\%)}& $5.11^{a}$\newline $7.03 ^{b}$ & $1.50 ^{ a}$\newline  $1.38 ^{b}$ & $4.33 ^{ a}$\newline $4.96 ^{b}$ & $1.03 ^{a}$\newline  $1.04 ^{b}$ \\
		\textit{Training Time (s)} & $205 ^{a}$\newline $248 ^{b}$ & $192 ^{ a}$\newline $234 ^{b}$ & $80 ^{ a}$\newline $105 ^{b}$ & $850 ^{a}$\newline $1055 ^{b}$ \\
		\textit{Forecasting time (ms)}  & $7.31 ^{S1}$\newline $735.1 ^{S2}$ & $7.65 ^{S1}$\newline  $757.36  ^{S2}$ & $1.04  ^{ S1}$\newline $105.42  ^{S2}$ & $16.8 ^{S1}$\newline $1712.8  ^{S2}$ \\
		\hline 
	\end{tabular*}
	$^a = DMA1$, $^b = DMA2$, $ S1 = $forecasting Scenario 1,  $ {S2} = $ forecasting Scenario 2.
\end{table}
\begin{table}[h!]
	\renewcommand{\arraystretch}{1.5}
	\caption{Comparison with the results of the previous publications}
	\label{Previous_RESULTS}
	\centering
	\begin{tabular*}{\linewidth}{p{3.5cm} p{3.2cm} p{2.0cm} p{1.5cm} p{1.5cm}}
		\hline
		Research &	Technology & Prediction scenario & MAPE & Complexity (AIC)\\ 
		\hline
		\cite{Antunes2018} & Combination of ANN, SVM, \newline$ k $-nearest-neighbors, and random forest & One-step daily WDF & $9.1\%$ & Not discussed\\
		\cite{Brentan2017} & Combination of SVM and Fourier method & One-step daily WDF & $3.41\%$ & Not discussed \\
		\cite{Mouatadid2017} & ELM & One-step daily WDF& $ 9.7\% $ & Not discussed \\
		\cite{Tiwari2016} & ELM with wavelet ANN & One-step daily WDF & $6.09\%$ & Not discussed\\
		\cite{Candelieri2017CSVM}  & Clustering with SVM & One-step hourly WDF & $ 5.9\% $& Not discussed\\
		\cite{Du2021LSTM}  & DWT and PCA with LSTM & One-step daily WDF & $ 1.83\% $& Not discussed\\
		\cite{Guancheng2018} & GRUN  & One-step quarterly WDF &$ 2.06\% $ & $ 179019$ \\
		\cite{Guancheng2018} & GRUN  & 96-step quarterly WDF & $ 4.33 \%$ & $ 179019$ \\
		\textbf{This research }& \textbf{GRU with \textit{k}-means Classification } & \textbf{One-step quarterly WDF }&\textbf{1.31\%} & \textbf{29396}\\
		\textbf{This research }& \textbf{GRU with \textit{k}-means Classification } & \textbf{96-step quarterly WDF }&\textbf{1.5\%} & \textbf{29396}\\
		\hline 
	\end{tabular*}
\end{table}
On the other hand, training the DCGRU model takes about 192 seconds and 234 seconds for DMA1 and DMA2, respectively. Training the GRUN model takes around 80 seconds and 105 seconds for DMA1 and DMA2, respectively, which means that the GRUN model is better than our model in terms of time complexity.

\begin{figure}[!h]
	\centering
	\includegraphics[scale=0.60]{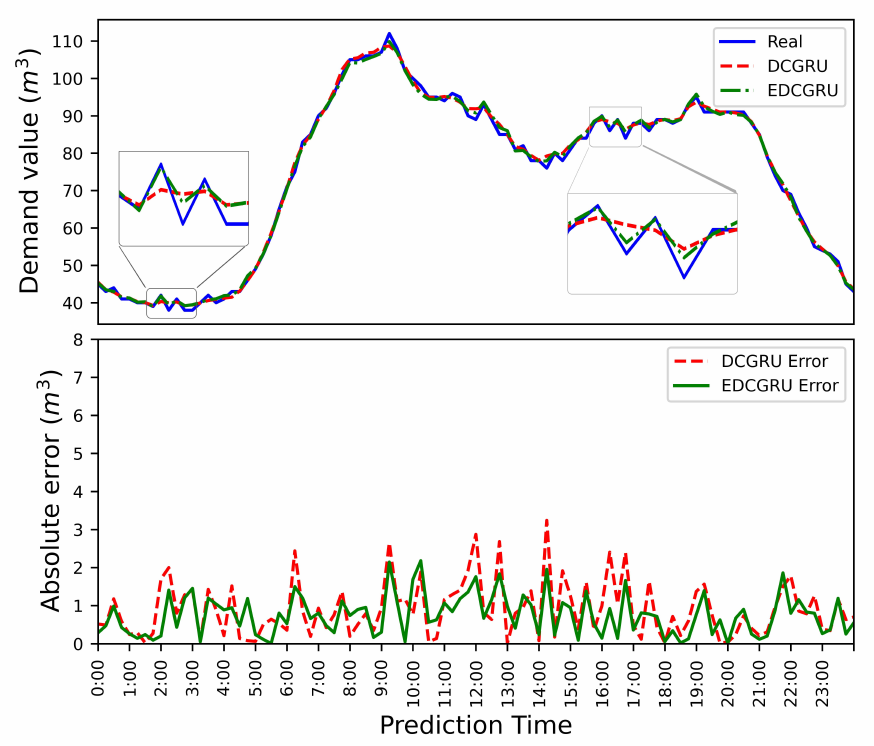}
	\caption{Prediction results and absolute errors for DMA1 in Scenario 1 before and after expanding the data.}
	\label{fig8}
\end{figure}
\begin{figure}[!h]
	\centering
	\includegraphics[scale=0.60]{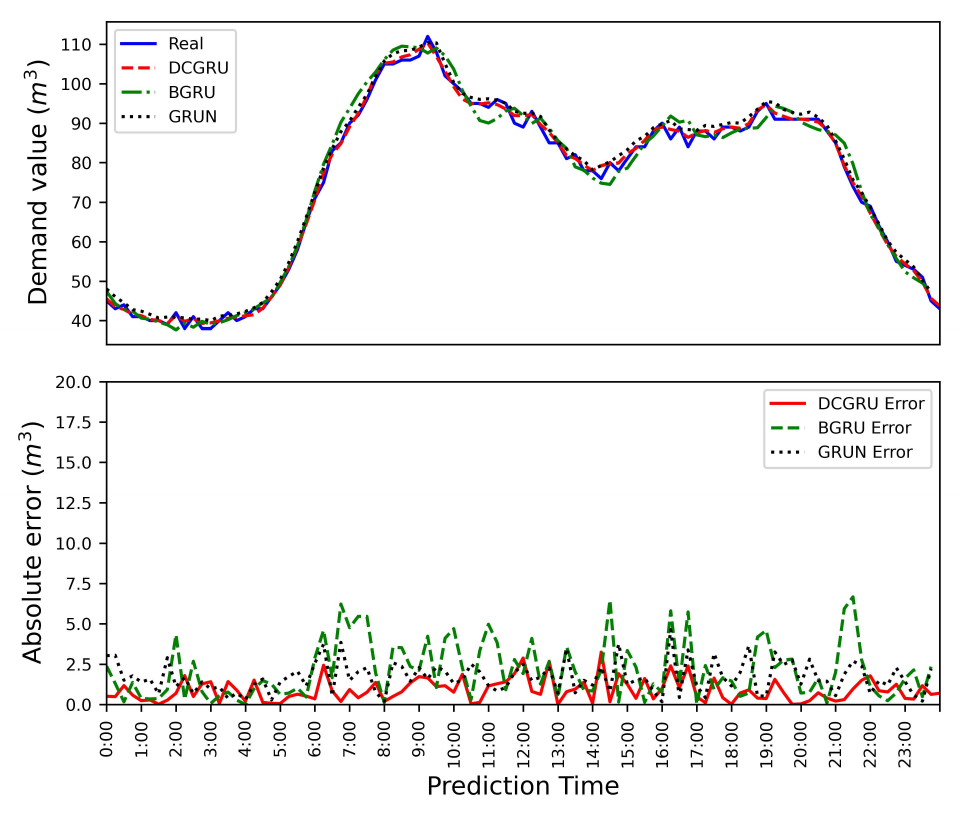}
	\caption{Prediction results and absolute errors for DMA2 in Scenario 1 before and after expanding the data.}
	\label{fig9}
\end{figure}
\begin{figure}[!h]
	\centering
	\includegraphics[scale=0.60]{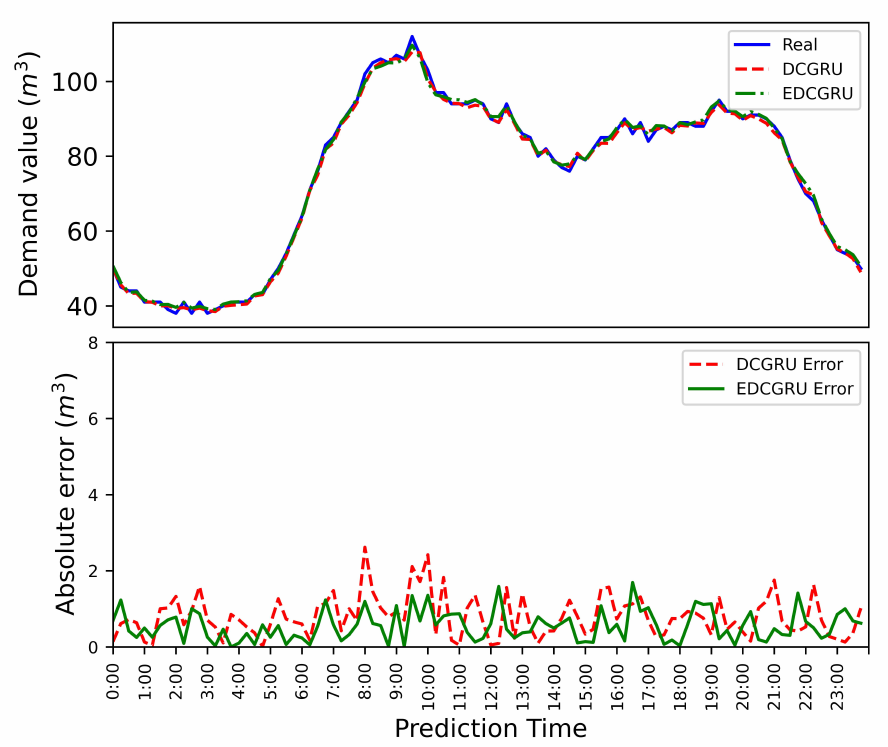}
	\caption{Prediction results and absolute errors for DMA1 in Scenario 2 before and after expanding the data.}
	\label{fig10}
\end{figure}
\begin{figure}[!h]
	\centering
	\includegraphics[scale=0.60]{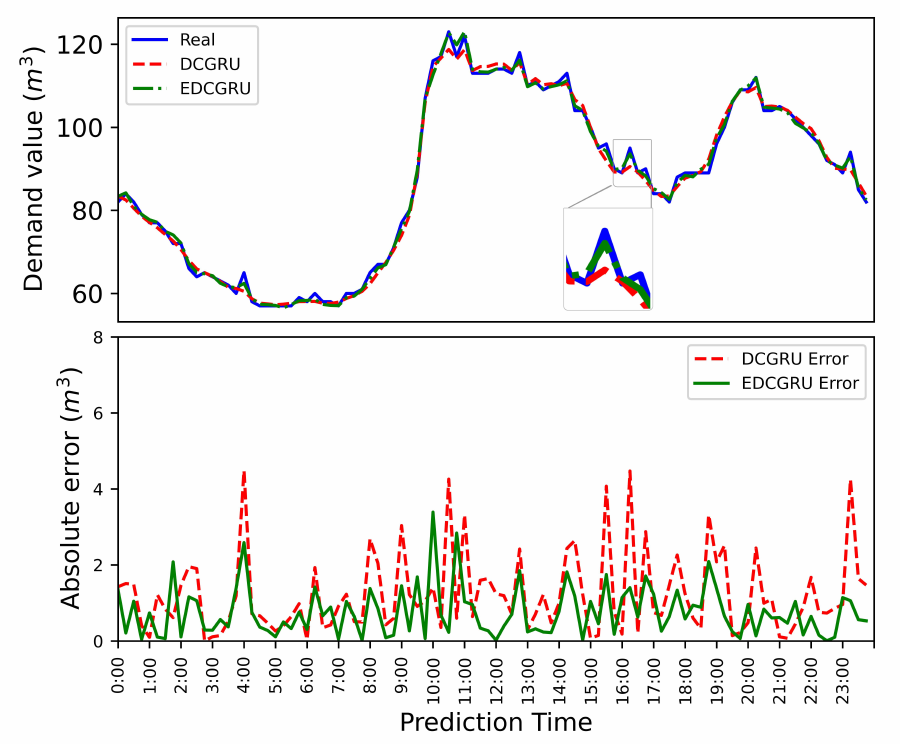}
	\caption{Prediction results and absolute errors for DMA2 in Scenario 2 before and after expanding the data.}
	\label{fig11}
\end{figure}

After expanding the data set, the forecasting accuracy becomes much better than before with the same number of parameters, making it the best in terms of space complexity. EDCGRU model has 9176 and 11873 of AIC for DMA1 and DMA2, and achieves 0.98\%, and 0.99\% of MAPE for DMA1 and DMA2, respectively, in Scenario 1, and 1.03\% and 1.04\% for DMA1 and DMA2, respectively, in Scenario 2. However, it has the worst time complexity, which basically caused by the large input size. Training the model needs 850 seconds and 1055 seconds for DMA1 and DMA2, respectively.

\section{Discussion}
\label{Sec5}
In this work, a new ANN structure for StWDF is investigated, aiming at minimising the number of the trainable parameters while maintaining a level of forecasting accuracy not less than that reported in the previous works. 

The proposed DCGRU model achieves this goal successfully, as shown in the results. The complexity of our model is reduced effectively compared to what achieved in the state-of-the-art. This is attributable to two factors; (i) using GRU cell instead of LSTM cell as done in some works in literature \citep{VanHoudt2020,Mu2020}. The GRU cell contains nine parameters as shown in equations \ref{Eq1}, \ref{Eq2}, and \ref{Eq3}. while the LSTM cell contains ten parameters \citep{ZHANG2020113226,xie2020motion}. (ii) Utilizing the classification to find the connection between water demand in different days devises a chance to handle this relationship via an ANN unit with fewer parameters, unlike the GRUN model where \cite{Guancheng2018} uses three GRU layers to handle the same relationship adding too many parameters to the model structure.
Although the BGRU model has the minimum number of parameters as illustrated in Table \ref{EVALUATION_RESULTS}, it can not reach a sufficient prediction accuracy in either of our scenarios, as a consequence, AIC of BGRU is significantly larger than that of DCGRU. Thanks again to the classification step, which enhances the DCGRU model accuracy by creating additional useful features so that our model achieves the best accuracy compared to the previous works with reasonable complexity, as the comparison results in Table \ref{Previous_RESULTS} clarify.

Furthermore, the created relationship between data from the same class plays an influential role in diminishing the influence of involving the predicted values in prediction in Scenario 2. Thus, no further modifications are required to solve the accumulative error problem. On the contrary, the GRUN model needs a correction model to solve this problem which adds many parameters and increase the complexity.

Computational load is influenced by the structure of the model as well as the input size. The larger the input size, the heavier the load. Although the design of the BGRU model is simple, it can not converge quickly.
Thus, more training epochs are required to arrive at the best accuracy, as shown in Table \ref{TRAINING_PARAMETERS}. This explains why BGRU has longer training time than DCGRU despite it has fewer parameters and both have the same input size.
On the other hand, GRUN has the fastest training process due to the tiny input size compared to the other three models, where it has input size of 15 records compared with 96 for DCGRU and BGRU, and 192 for EDCGRU models. Although the use of full-day water demand measurements as an input for the DCGRU model increases the computational time, it guarantees good performance in Scenario 2.

The EDCGRU model has the lowest MAE for both DMAs in both scenarios, which emphasizes that expanding the data impacts the overall accuracy by increasing the linearity between the sequential data. However, this expansion reflects negatively on the computational load because of increment in the input size. Thus both training and forecasting time dramatically rise.

The training time differs based on the DMA because every DMA has a different pattern within the data which may take different time to approximate. In contrast, training time does not change when forecasting scenario changes because each model is trained once and used for both scenarios.
On the other hand, forecasting time depends only on the forecasting process, making it change based on the forecasting scenario and remain the same for different DMAs.
\section{Significance of results} 
\label{Sec6}
The prediction models proposed in this research provide accurate forecasting of water demand, where the MAE of both CDGRU and ECDGRU does not exceed 1 $ m^3/15 $ minutes in Scenario 1, and 1.2 $m^3/15 $ minutes is Scenario 2. Such an accuracy guarantees accurate pumping, which in turn secures a satisfactory service while reducing the risk of sabotaging the pipes by high pressure and reduce the cost of maintenance. 

Additionally, water demand data are confidential data because they are rich with private information. Thus training a machine learning model on data sets of different areas requires following a training strategy that respects this privacy. This strategy could be federated learning  \citep{li2020federated}. Federated learning is an approach where the model is trained on different data sets located in different databases without moving the data; thus, several transitions are required to complete the training. To ease the transitions, a small size model is preferred. Knowledge transformation is another practical approach that aims to reduce training time using a DL model already trained on similar data and retrain it or retrain part of it on the required date. This approach also requires transferring the trained model with its parameters to the target data set. This work result in a DL model that is at least six times smaller than what proposed in the literature for StWDF.

Scientifically, the method proposed to enhance the accuracy at the extreme points can be applied for any other time series problem to reduce the nonlinearity of the data. In fact, the majority of data series such as wind speed data, pollution rate data, and daily temperature suffer from high nonlinearity. Adding virtual data in between the actual data in the way proposed in this research raises the linearity between data, which lower the prediction error.
However, it still needs more investigation regarding the training time, where increasing the data size results in higher training time to reach an acceptable accuracy.
 \section{Conclusion} 
 \label{SecConc}
This research investigates a novel DL model for StWDF and proposes a novel strategy for mitigating the error at the extreme points. The main goal of the new design is to minimize the place complexity of the model while keeping high accuracy levels in two forecasting scenarios, one-step and multi-step forecasting scenarios, in addition to reducing the accumulative error problem in multi-step forecasting scenario.
 
To enhance the performance of the model, the historical data of water demand is classified into four classes. Such a step creates more relationships between data to enhance the prediction accuracy, and relieves the reliance on the sequential relationship and minimizes the accumulative error. Moreover, this prior clustering process reforms the data shape such that a simple ANN model can approximate additional characteristic of the data.
 
A comparison between the proposed model and the BGRU model exposes the benefit of the classification step to lower the error. While a comparison with state-of-the-art shows that the proposed design guarantees a remarkable improvement, where the complexity of the model is reduced 6 times of what achieved in the literature while preserving the best prediction accuracy, 98:69\% and 98:5\% for Scenario 1 and Scenario 2, respectively. On the other hand, expanding the data set adding virtual values within the data to reduce the nonlinearity at extreme points is proved to be effective and reliable. It is found that adding more virtual values within the data does not necessarily lead to better accuracy at  extreme points, this is due to the accumulative error that grows when relying on more predicted virtual values to forecast one actual value. This work found that one virtual value can reduce the error at extreme points by 30\% of its original value.
 
Nevertheless, the model presented in this work relies on water demand history, making it vulnerable to abnormalities in water demand. This problem will be considered in future work, in addition to optimizing the model structure and the training parameters based on an evolutionary method.
 
\section*{Acknowledgment}
The authors would like to acknowledge the helpful comments of Prof. Shuming Liu of Tsinghua University, China throughout our research and sharing their data set with us.

This work was supported by the National Natural Science Foundation of China under Grant  62073031, 62061160371,  and the Fundamental Research Funds for the China Central Universities of USTB under Grant FRF-TP-19-001C2.
\section*{Declaration of Competing Interest}
The authors declare that they have no known competing financial interests or personal relationships that could have appeared to influence the work reported in this paper.
\section*{CRediT authorship contribution statement}
\textbf{Tony Salloom:} Conceptualization, Formal analysis, Investigation, Software, Writing – original draft. \textbf{Okyay Kaynak:} Data curation, Methodology, Validation, Writing – review \& editing, Supervision. \textbf{Wei He:} Funding acquisition, Writing - review \& editing, Resources, Supervision
\bibliographystyle{cas-model2-names}
\bibliography{References.bib}
\newpage
\begin{wrapfigure}{l}{0.15\textwidth}
	\centering
	\includegraphics[width=0.17\textwidth]{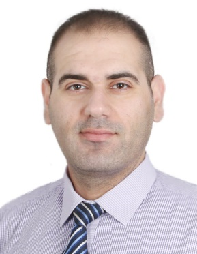}
\end{wrapfigure}
\noindent\textbf{Tony Salloom}
received the B.Sc. in electronics and computer engineering from Aleppo University, Aleppo, Syria, in 2008, and the M.Eng. in information and telecommunication engineering from the University of Science and Technology Beijing, Beijing, China, in 2016. \\
From 2009 to 2013, he served as a Database Engineer at the Syrian Telecommunication Company, Aleppo, Syria. He is currently pursuing the Ph.D. degree with the School of Automation and Electrical Engineering, University of Science and Technology Beijing, Beijing, China. His current research interests include Deep learning, intelligent control systems, and robotics.
\newline

\begin{wrapfigure}{l}{0.15\textwidth}
	\centering
	\vspace{-14pt}
	\includegraphics[width=0.17\textwidth]{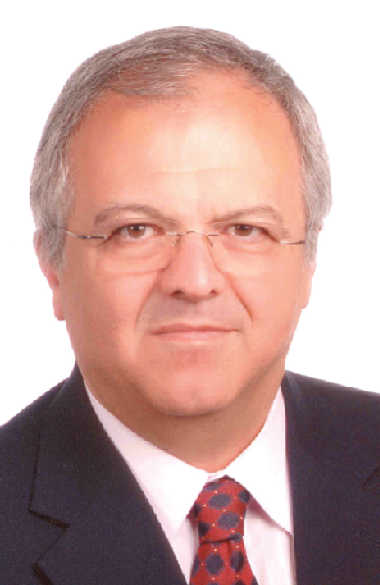}
	\vspace{-19pt}
\end{wrapfigure}
\noindent\textbf{Okyay Kaynak}
received the B.Sc. degree with first class honors and Ph.D. degrees in electronic and electrical engineering from the University of Birmingham, UK, in 1969 and 1972 respectively. From 1972 to 1979, he held various positions within the industry. In 1979, he joined the Department of Electrical and Electronics Engineering, Bogazici University, Istanbul, Turkey, where he is currently a Professor Emeritus, holding the UNESCO Chair on Mechatronics. He is also a 1000 People Plan Professor at University of Science \& Technology Beijing, China. He has hold long-term (near to or more than a year) Visiting Professor/Scholar positions at various institutions in Japan, Germany, U.S., Singapore and China. His current research interests are in the fields of intelligent control and mechatronics. He has authored three books, edited five and authored or co-authored more than 450 papers that have appeared in various journals and conference proceedings.

Dr. Kaynak has served as the Editor in Chief of \emph{IEEE Trans. on Industrial Informatics} and \emph{IEEE/ASME Trans. on Mechatronics} as well as Co-Editor in Chief of \emph{IEEE Trans. on Industrial Electronics}. Additionally, he is on the Editorial or Advisory Boards of a number of scholarly journals. In 2016, He received the Chinese Governments Friendship Award and Humboldt Research Prize. Most recently he was awarded the Academy Price of Turkish Academy of Sciences (2020).

Dr. Kaynak is active in international organizations, has served on many committees of IEEE and was the president of IEEE Industrial Electronics Society during 2002-2003. He was elevated to IEEE fellow status in 2003.
\newline

\begin{wrapfigure}{l}{0.15\textwidth}
	\vspace{-14pt}
	\includegraphics[width=0.17\textwidth]{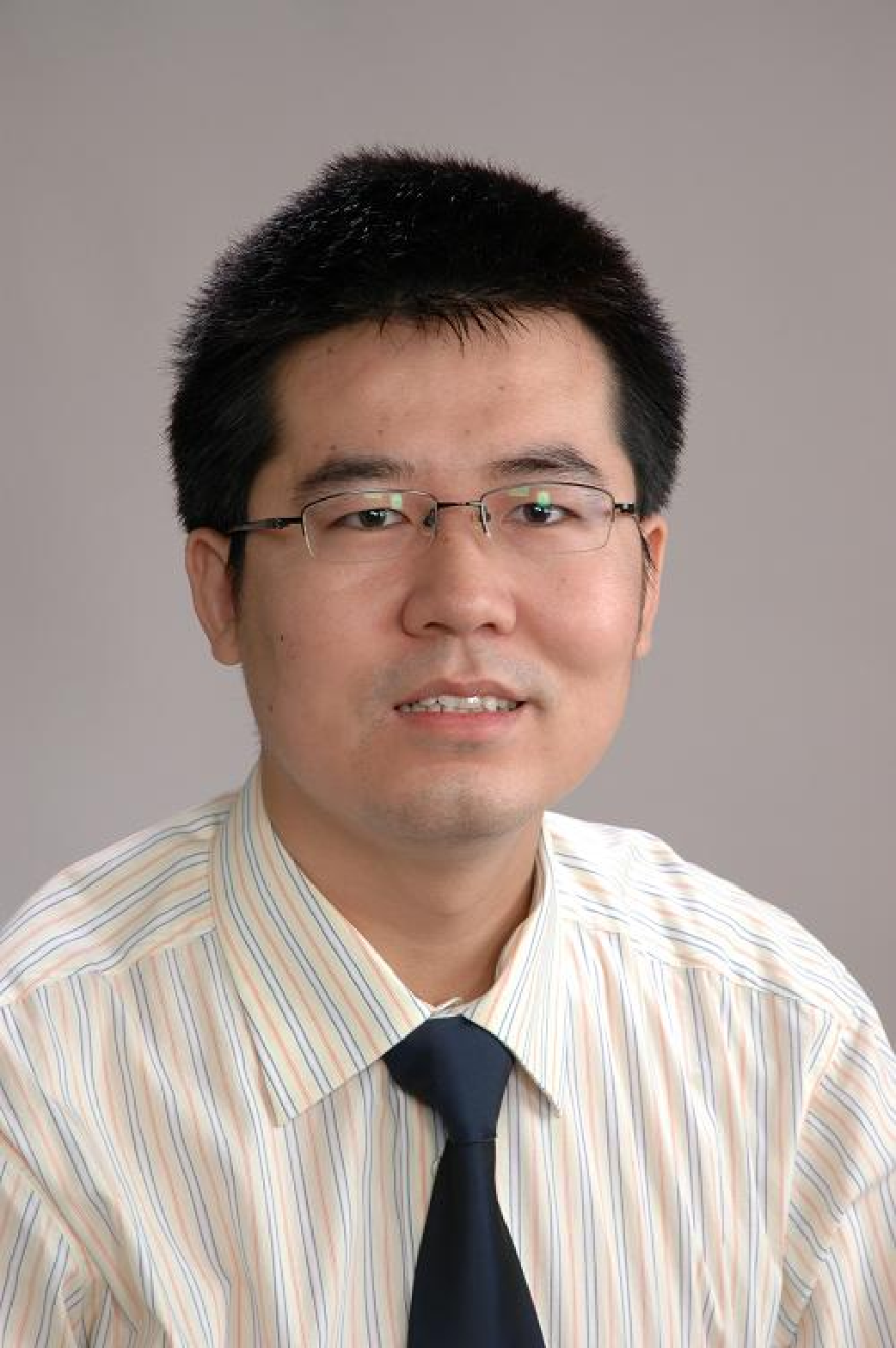}
	\vspace{-19pt}
\end{wrapfigure}
\noindent\textbf{Wei He}
(S'09-M'12-SM'16) received his B.Eng. in automation and his M.Eng. degrees in control
science and engineering from College of Automation Science and Engineering, South China University of Technology (SCUT), China, in 2006 and 2008, respectively, and his Ph.D. degree in control
science and engineering from from Department of Electrical \& Computer Engineering, the National University of Singapore (NUS), Singapore, in 2011.

He is currently working as a full professor in School of Automation and Electrical Engineering, University of Science and Technology Beijing, Beijing, China. He has co-authored 2 books published in Springer and published over 100 international journal and conference papers. He was awarded a Newton Advanced Fellowship from the Royal Society, UK in 2017. He was a recipient of the IEEE SMC Society Andrew P. Sage Best Transactions Paper Award in 2017. He is serving the Chair of IEEE SMC Society Beijing Capital Region Chapter. He is serving as an Associate Editor of \emph{IEEE Transactions on Robotics}, \emph{IEEE Transactions on Neural Networks and Learning Systems}, \emph{IEEE Transactions on Control Systems Technology}, \emph{IEEE Transactions on Systems, Man, and Cybernetics: Systems}, \emph{IEEE/CAA Journal of Automatica Sinica}, \emph{Neurocomputing}, and an Editor of \emph{Journal of Intelligent \& Robotic Systems}.
His current research interests include robotics, distributed parameter systems and intelligent control systems.
\end{document}